%% file: root.tex
\documentclass[letterpaper, 10pt, conference]{IEEEtran}

\IEEEoverridecommandlockouts                              

\input{Macro}

\title{\LARGE \bf
GainAdaptor: Learning Quadrupedal Locomotion with Dual Actors for Adaptable and Energy-Efficient Walking on Various Terrains
}

\author{
Mincheol Kim$^{1,*}$, Nahyun Kwon$^{1,*}$, Jung-Yup Kim$^{1,\dagger}$
\thanks{This work was supported by the Ministry of Trade, Industry and Energy (MOTIE, Korea) under the Industrial Technology Innovation Program. Grant No. 20026194, "Development of Human-Life Detection and Fire-Suppression Solutions based on Quadruped Robots for Firefighting and Demonstration of Firefighting Robots and Sensors". * These authors contributed equally to this work. ${\dagger}$ Corresponding author. $^{1}$M. Kim, N. Kwon, and J. Kim are with the Humanoid Robot Research Laboratory, Department of Mechanical Design and Robot Engineering, Seoul National University of Science and Technology, Seoul, 01811, Republic of Korea ({\tt\small kmc96, kwonnahyun, jyk76}@seoultech.ac.kr).}%
}

\begin{document}

\maketitle
\thispagestyle{empty}
\pagestyle{empty}

\begin{abstract}

Deep reinforcement learning (DRL) has emerged as an innovative solution for controlling legged robots in challenging environments using minimalist architectures. Traditional control methods for legged robots, such as inverse dynamics, either directly manage joint torques or use proportional-derivative (PD) controllers to regulate joint positions at a higher level. In case of DRL, direct torque control presents significant challenges, leading to a preference for joint position control. However, this approach necessitates careful adjustment of joint PD gains, which can limit both adaptability and efficiency. In this paper, we propose GainAdaptor, an adaptive gain control framework that autonomously tunes joint PD gains to enhance terrain adaptability and energy efficiency. The framework employs a dual-actor algorithm to dynamically adjust the PD gains based on varying ground conditions. By utilizing a divided action space, GainAdaptor efficiently learns stable and energy-efficient locomotion. We validate the effectiveness of the proposed method through experiments conducted on a Unitree Go1 robot, demonstrating improved locomotion performance across diverse terrains. Supplementary videos are available at https://sites.google.com/view/gainadaptor
\end{abstract}

\begin{IEEEkeywords}
Dual actors; Deep reinforcement learning; Quadrupedal locomotion; Gain optimization
\end{IEEEkeywords}

\input{Sections/01_Introduction}

\input{Sections/02_Related_works}

\input{Sections/03_Proposed_Method}

\input{Sections/04_Experimental_Results}
\input{Sections/05_Conclusion}


{\small
\bibliographystyle{unsrt}
\bibliography{egbib}
}

                                  %

\end{document}

%% file: Macro.tex
\usepackage{times}
\usepackage{epsfig}
\usepackage{color} 
\usepackage{amsmath}
\usepackage{amssymb}
\usepackage{graphicx}
\usepackage{multirow}
\usepackage{tabularx}
\usepackage{adjustbox}
\usepackage{siunitx}
\usepackage{textcomp}
\usepackage{array}
\usepackage{pifont}
\usepackage{booktabs}
\usepackage{amsfonts}
\usepackage[ruled]{algorithm2e}
\usepackage{hyperref}
\usepackage{cleveref}
\usepackage{verbatim}

\usepackage{authblk}

\usepackage[letterpaper, top=57pt, bottom=43pt, left=48pt, right=48pt]{geometry}

\crefname{table}{Table}{Tables}
\Crefname{table}{Table}{Tables}
\usepackage{stfloats}
\usepackage{cite}
\usepackage{stmaryrd} 
\usepackage{latexsym} 
\usepackage{url}
\usepackage{subcaption} 
\usepackage{booktabs}
\usepackage{algpseudocode}
\usepackage{kotex}

\newcommand{\figref}[1]{Fig.~\ref{#1}}

\newcommand{\secref}[1]{Sec.~\ref{#1}}

\newcommand{\Tabref}[1]{Table~\ref{#1}}

\newcommand{\etal}{\textit{et al.}}

\usepackage{xcolor,colortbl}
\definecolor{Gray1}{gray}{0.85}
\definecolor{Gray2}{gray}{0.65}

\usepackage{pifont}

%% file: Sections/01_Introduction.tex
\section{Introduction}
\label{sec:intro}

Over the past decade, deep reinforcement learning (DRL) has gained significant attention as an innovative approach to robot control tasks. Remarkable progress has been made in enabling robots to perform a wide variety of tasks across diverse environments using relatively simple control structures, minimizing the complexity of robot actuation and decision-making processes. These advancements have enabled robots to navigate challenging terrains such as mountainous areas, beaches, and deformable surfaces like sand, all while maintaining mobility and stability~\cite{ref1, eth_mountain, deformable}.

Despite these advancements, several challenges remain, particularly related to overcoming battery capacity limitations and extending overall operational duration~\cite{battery, chen2017energy, valsecchi2024accurate}. These limitations directly affect the robot’s ability to complete long missions in demanding environments. Extending a robot's operating time by restricting joint torque or speed can mitigate these issues, however such constraints severely compromise adaptability, particularly in complex environments. Therefore, innovative solutions are required to simultaneously enhance both energy efficiency and environmental adaptability.

Achieving these objectives using DRL requires careful consideration of action space design. Typically, there are two main approaches to controlling the locomotion of a legged robot: direct control of low-level joint torque~\cite{lillicrap2015continuous, schulman2015high, chen2023learning} or reference joint position control through a proportional-derivative (PD) controller~\cite{margolis2023walk, ji2022concurrent, dreamwaq}.

\input{Images/01_Teaser}
Directly generating torque for each joint motor can enhance the robot’s agility and optimize its movements. However, this method has significant drawbacks: it requires extended learning periods and precise coordination among the joints to achieve effective outcomes.

Alternatively, reference joint position control through PD controller allows the robot to calculate target joint positions by combining desired motions with its initial home posture. The PD controllers then use these target positions to generate the required joint torques. This approach, in contrast to direct torque control, is less sensitive to the robot’s hardware characteristics and significantly shortens learning time by providing a stable reference posture during training. As a result, many studies have favored using reference joint positions for locomotion inputs~\cite{ref1, eth_mountain, deformable, margolis2023walk, ji2022concurrent, dreamwaq}.

However, this approach has its limitations. Manually adjusting the joint PD gains to appropriate levels often requires human intervention, which can be impractical in real-time or in rapidly changing environments. Techniques such as domain randomization and gain tuning help robots adapt to various terrains, however, they often confine performance by limiting the allowable range of the PD gains. A more adaptive control system is required to overcome these limitations by dynamically adjusting to different environments without human intervention.

To address these challenges, we propose {\bf GainAdaptor}, an adaptive gain control framework for quadrupedal locomotion on diverse terrains. This framework introduces a novel dual-actor algorithm that dynamically adjusts the joint PD gains in response to changing ground conditions, thereby enhancing the robot’s adaptability while reducing power consumption and extending its operational time.

GainAdaptor also integrates a terrain state estimator, which helps determine the appropriate locomotion strategy for different terrains. By leveraging this estimator, GainAdaptor not only improves the robot’s ability to navigate complex environments but also addresses the critical issue of energy efficiency.

To validate the effectiveness of the proposed method, we conducted a series of real-world experiments and practical demonstrations using a Unitree Go1 robot, as shown in \figref{fig:teaser}. The experiments focused on diverse terrains to thoroughly test the adaptability and energy efficiency improvements provided by GainAdaptor.

In summary, our contribution is threefold:
\begin{enumerate}
\item
{\bf GainAdaptor Framework:} We propose GainAdaptor, an adaptive gain control framework that enhances adaptability across various terrains while improving energy efficiency. The framework utilizes a novel dual-actor algorithm to dynamically adjust the joint PD gains in response to changing ground conditions.

\item
{\bf Terrain Classifier:} By incorporating a terrain state classifier within GainAdaptor, we enable the robot to identify and implement the most suitable locomotion strategy for different terrains, further improving performance in diverse environments.

\item
{\bf Efficient and Robust Performance:} Through extensive real-world experiments, including a battery efficiency test, we demonstrate that GainAdaptor enables robust locomotion across diverse terrains while significantly reducing torque and power consumption. This results in enhanced energy efficiency, extended operational endurance, and a tangible improvement in battery performance.
\end{enumerate}

%% file: Images/01_Teaser.tex
\begin{figure}[t]
\centering 
\begin{tabular}{c}
\includegraphics[width=0.99\linewidth]{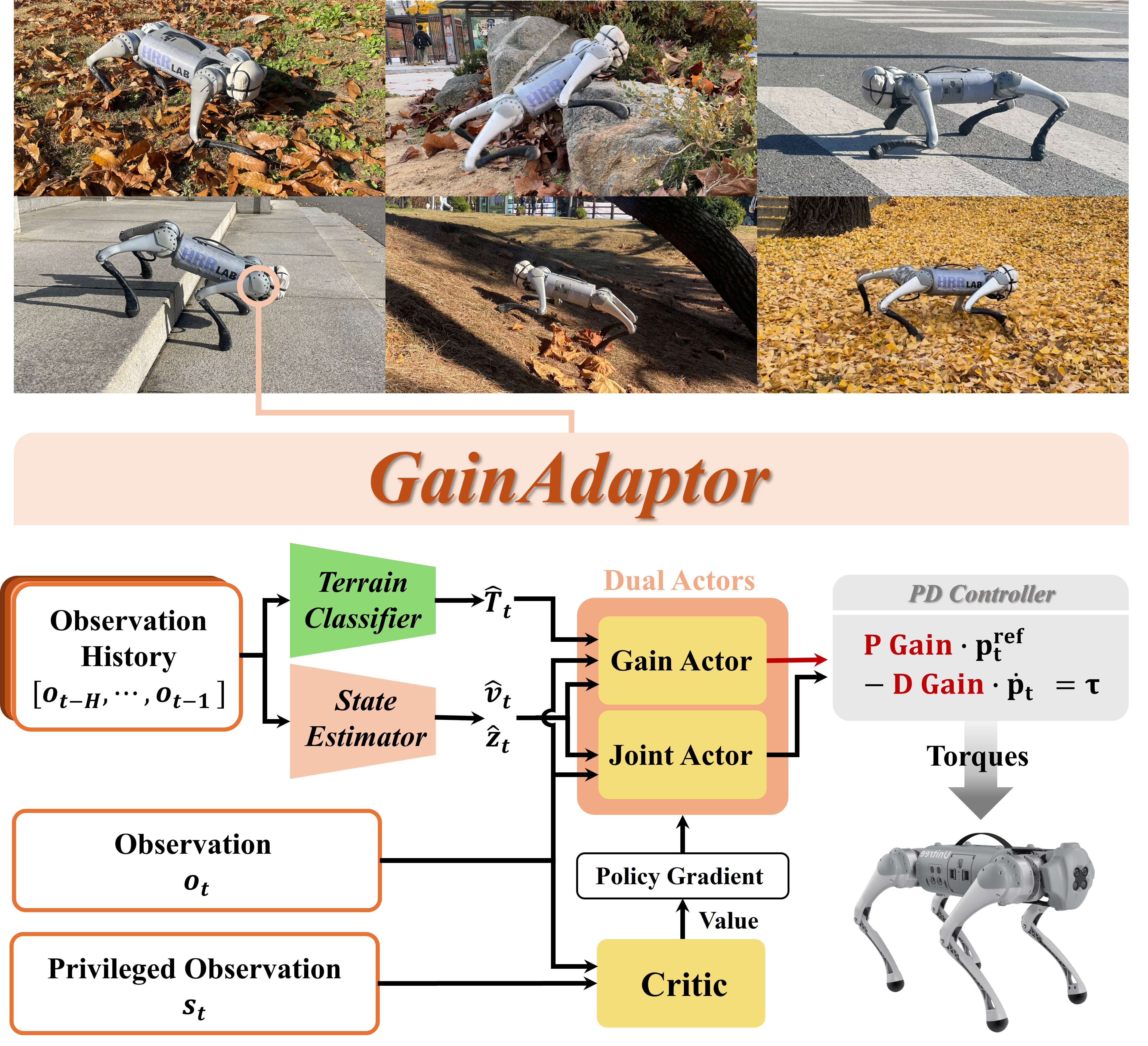}  \\ 
\end{tabular}

\caption{{\bf Adaptive quadrupedal locomotion with optimized joint PD gains.}
The proposed GainAdaptor framework enables quadrupedal robots to dynamically adjust joint PD gains in response to changing environments, resulting in energy-efficient locomotion. This framework is demonstrated using a Unitree Go1 robot.
}
\vspace{-0.1in}
\label{fig:teaser} 
\end{figure}

%% file: Sections/02_Related_works.tex
\section{Related Works}
\label{sec:related works}

\subsection{Deep Reinforcement Learning}

Deep reinforcement learning (DRL) has emerged as a promising alternative for designing robot controllers without requiring the extensive mathematical knowledge typically needed for traditional model-based controllers such as Model Predictive Control (MPC) and Whole Body Control (WBC)\cite{mpc,wbc}. These traditional controllers rely on a complex mathematical foundation, incorporating optimization, kinematic and dynamic models along with state feedback control. In contrast, DRL enables learning-based control without requiring explicit model derivation.

Recent research has primarily focused on reducing training times\cite{ref10}, achieving agile locomotion, and bridging the sim-to-real gap. For instance, the work by Rudin \etal\cite{ref10} successfully reduced reinforcement learning training time by leveraging Nvidia's Isaac Gym\cite{makoviychuk2021isaac} and parallel GPU processing. Similarly, the study by Nahrendra \etal\cite{dreamwaq} has made significant progress in enabling legged robots to perform agile locomotion across variable terrains without relying on visual inputs. Other notable advancements include robot running at high speed\cite{rapid}, executing parkour-style maneuvers using visual feedback\cite{zhuang2023robot,cheng2023extreme}, and navigating deformable surfaces like sand\cite{deformable}.

In addition to agility, research efforts have focused on making legged robots more robust, even in the event of motor failures\cite{Kim2024LearningQL}, while adapting to changing environmental states. Domain randomization techniques\cite{domain1,domain2} are commonly employed to bridge the sim-to-real gap, and teacher-student training methodologies\cite{rma,rapid} have been utilized to facilitate generalization across different terrains.

However, previous studies have primarily used reference joint positions as the action space in DRL while fixing P and D gains at low values. Since the joint torque is ultimately generated by the PD controller, real-time optimization of P and D gains is crucial for enhancing locomotion performance and energy efficiency, especially in diverse environments.

\subsection{Gain Tuning and Optimization}

PID controllers are widely used across various applications\cite{bennett1993development}, with gain adjustment being crucial for optimal performance. Traditional methods like trial-and-error and Ziegler-Nichols\cite{Ziegler,Ziegler2} are commonly employed. However, most existing studies focus on finding and fixing optimized gains ahead of time, which limits the dynamic adjustment of PID gains in response to changing environments. This limitation becomes especially problematic in legged robots that must adapt to different terrains. Once-set PID gains may not be suitable for all environmental conditions.

Recent studies have highlighted the challenges of optimizing gains dynamically. Xie \etal\cite{p_gain&random} found that high P gain can cause instability on uneven terrains due to reduced compliance, while low P gain acts like torque controllers, providing stability but increasing joint position tracking errors. Similarly, Smith \etal\cite{smith2022walk} demonstrated that low D gain can lead to joint oscillations and instability, while high D gain slows the robot’s response to changes in target joint position.

To address these issues, Siekmann \etal\cite{siekmann2021sim} incorporated PD gains into the DRL action space for optimization. However, allocating too many parameters to the action space degraded learning performance due to increased complexity.

Our proposed method resolves this issue by dividing the action space and employing a dual-actor structure. This approach allows for stable and efficient dynamic optimization of the PD control system.

%% file: Sections/03_Proposed_Method.tex
\section{Method: GainAdaptor Framework}
\label{sec:method}

\subsection{Learning Environment of Dual Actors}
\label{sec:Dual}
\subsubsection{Dual Actors}
The primary motivation for employing a dual-actor system is to separate the tasks of PD gain adjustment and locomotion control. In the traditional single-actor-critic approach, a single actor must manage both tasks simultaneously, which can lead to conflicts due to differing objectives and action output scales. This often results in suboptimal learning performance. By using two specialized actors, our system improves the robot's adaptability to changing environments while ensuring stable and efficient control.

The gain actor dynamically tunes the proportional (P) and derivative (D) gains of the robot's joints, allowing it to adapt stiffness and flexibility to the terrain. Here, $P_{init}$ and $D_{init}$ gain are initialized at 28 and 0.7. This ensures stable and responsive movement across diverse environments, achieving smooth and efficient locomotion.

The joint actor utilizes the optimized PD gains to control the robot's physical movements. It generates the reference positions of each joint $q_{\text{des}}$, calculated using the initial default position $q_{\text{def}}$ and scaled action outputs from the actor. This enables the robot to execute suitable locomotion strategies for varying terrains. By applying movement policies learned from various environments, this actor maintains the robot’s stability and energy efficiency during operation.

As illustrated in~\figref{fig:teaser}, the interaction between the dual actors (\secref{sec:Dual}), along with the observation and reward settings (\secref{sec:Setting}) and the overall learning structure (\secref{sec:terrain}), enhances the robot’s adaptability, stability, and locomotion efficiency.

\subsubsection{Divided Action Space}

The GainAdaptor framework operates within a shared environment, dividing the action space between the two actors. The action space is defined as $a_t\in\mathbb{R}^{36}$, with the first 12 actions ($a_t^{\text{pos}}\in\mathbb{R}^{12}$) representing reference joint positions and the remaining 24 actions ($a_t^{\text{PD}}\in\mathbb{R}^{24}$) corresponding to P and D gains.

Inspired by existing joint position control methods, we accelerate learning by incrementally adjusting desired joint angles based on the robot's initial posture. Each joint’s default position $q_{\text{def}}$ is set to the robot’s initial posture and the desired reference position $q_{\text{des}}$ is computed as:
\begin{equation} q_{\text{des}} = q_{\text{def}} + \alpha_{\text{pos}} a_{t}^{\text{pos}}
\label{Equ
} 
\end{equation}
where $a_{t}^{\text{pos}}$ represents the action for the reference joint position, and $\alpha_{\text{pos}}$ is a scaling factor set to 0.25.

The desired joint torque $\tau_{des}$ is then calculated using a modified PD control equation: 
\begin{equation} \tau_{\text{des}} = \left( P_{init} + \alpha_{\text{gain}} a_{t}^{\text{P}} \right) (q_{\text{des}} - q_{t}) - \left( D_{init} + \alpha_{\text{gain}} a_{t}^{\text{D}} \right) \dot{q}_{t}
\end{equation} 
where $q_t$ and $\dot{q}_t$ represent the current joint position and joint angular velocity, respectively. $a_{t}^{\text{P}}$ and $a_{t}^{\text{D}}$ are actions for PD gain adjustment, and $\alpha_{\text{gain}}$ is a scaling factor set to 0.5 to prevent excessive changes in the gain values.

\subsection{Training Setup}
\label{sec:Setting}
\subsubsection{Training data acquisition}

Training data is categorized into two types: observation $o_t$, which includes real-world measurable data for moving the robot, and privileged observation $s_t$, which is available only in simulation. Observation $o_t \in \mathbb{R}^{66}$, shared across the critic and actors, includes gravity vector $g_t \in \mathbb{R}^{3}$, joint angle $q_t \in \mathbb{R}^{12}$, joint angular velocity $\dot{q}t \in \mathbb{R}^{12}$, body velocities in the $x$ and $y$ axes, yaw angular velocity of the body, command value $c_t \in \mathbb{R}^{3}$, and the last action $a_{t-1} \in \mathbb{R}^{36}$.

Privileged observation $s_t \in \mathbb{R}^{197}$, used only by the critic, includes linear and angular base velocities $(v_t, \omega_t)\in\mathbb{R}^{6}$, terrain height scan $h_t \in \mathbb{R}^{187}$, and terrain type $\hat{T}_t \in \mathbb{R}^{4}$, representing terrain conditions such as level ground, slopes, and stairs.
\input{Tables/01_Reward_table}

\subsubsection{Reward}

We extended a baseline reward function from prior studies \cite{ji2022concurrent, dreamwaq} to incorporate joint PD gains as an action, directly influencing the PD controller’s joint position estimation. This addition significantly enhanced locomotion adaptability in real-world environments.

However, integrating the gain actor introduced instability in the torque computed by the PD controller. During the early stages of training, we observed that excessively low positive rewards hindered effective learning. To address this, we slightly increased the weight of the linear velocity tracking reward, a positive reward component, ensuring stable and efficient training progression.

Additionally, the P gain tended to converge to excessively low values. To address this, a limiting reward was introduced to maintain the P gain above a practical threshold, ensuring stability in the simulation-to-reality transfer. The full reward structure is summarized in~\Tabref{table:Reward_table}.

\subsection{Architecture}
\label{sec:terrain}
\subsubsection{Asymmetric Dual-Actor Structure}

Building upon prior studies that demonstrated the effectiveness of actor-critic algorithms in utilizing privileged observations $s_t$ through the interaction of actor and critic networks\cite{pinto2017asymmetric}, we adopted an asymmetric dual-actor structure in the GainAdaptor framework, inspired by the methodology in \cite{dreamwaq}. This structure enables the robot to efficiently optimize both joint position control and adaptive PD gain adjustment by utilizing two specialized actors with distinct roles.

Both actors share latent vectors $\hat{z}_t$ and observations $o_t$ as inputs, while the gain actor additionally leverages terrain surface information $\hat{T}_t$ to dynamically adapt PD gains. Entropy and log-probability for each actor are calculated independently to facilitate effective learning and optimization. For instance, the entropy and log probability for the joint actor’s probability distribution $p_i$ and the gain actor’s probability distribution $q_j$ are computed as:

\begin{equation}
\label{equ:log}
\text{Log Probability} = \sum_{i=1}^{12} \log p_i + \sum_{j=13}^{36} \log q_j \end{equation}

\begin{equation}
\label{equ:entorpy}
\text{Entropy} = - \sum_{i=1}^{12} p_i \log p_i - \sum_{j=13}^{36} q_j \log q_j \end{equation}

By jointly optimizing locomotion control and adaptive gain adjustment, the system achieves a balanced trade-off between the two, enabling efficient and robust locomotion across diverse terrains. The dual-actor full system is trained using the Proximal Policy Optimization (PPO) algorithm \cite{schulman2017proximal}. The optimization process focuses on maximizing the total expected reward, defined as:

\begin{equation}
\label{equ:teacher_objective}
J(\pi) = \mathbb{E}_{r \sim p(r|\pi)} \left[ \sum_{t=0}^{T-1} \gamma^t r_t \right] \end{equation}

where $\gamma$ is the discount coefficient, $r_t$ is the reward at time $t$, and $T$ is the length of the scenario.  This dual-actor approach simultaneously enhances terrain adaptability and energy efficiency, enabling smooth adaptation to diverse environments.

\subsubsection{Estimator and Classifier Networks}

To enhance adaptability and stability in quadrupedal locomotion across diverse terrains, we designed two independent neural networks: the Terrain Classifier and the State Estimator. These networks provide complementary information and serve as core components of the GainAdaptor framework.

The Terrain Classifier processes the observation history $o^H_t$ to predict the current terrain type $\hat{T}_t$. It utilizes a multi-class classification model to identify four terrain types: level ground, slope, rough terrain, and stairs. The network is trained using cross-entropy loss, defined as:

\begin{equation} 
L_{CrossEntropy} = - \sum_{i=1}^{4} T_i \log(\hat{T}_i)
\label{equ:terrain_loss}
\end{equation}

where $T_i$ represents the ground truth terrain class, and $\hat{T}_t$ is the predicted probability for class $i$. This classifier provides terrain information critical for the robot's gain adaptation.

The State Estimator also takes $o^H_t$ as input and estimates both the robot's velocity $\hat{v}_t$ and its latent state $\hat{z}_t$. Velocity estimation and latent state modeling are pivotal for achieving stable locomotion, particularly for blind robots. Previous studies have demonstrated the efficacy of latent variable $z_t$ in implicitly encoding terrain information without relying on vision sensors \cite{dreamwaq, ji2022concurrent}.

Inspired by these approaches, we implemented a $\beta$-variant autoencoder ($\beta$-VAE) mechanism. The $\beta$-VAE network leverages Kullback-Leibler (KL) divergence\cite{kullback1951information} for latent state regularization and mean squared error (MSE) loss for velocity estimation. The total loss function is expressed as:

\begin{equation} 
L_{VAE} =  MSE(\hat{v_t}, v_t) + \beta D_{KL} \left( q(z_t \mid o_{t}^H) \parallel p(z_t) \right)
\label{equ:VAELOSS}
\end{equation}

where $q(z_t \mid o_{t}^H)$ denotes the posterior distribution of the latent state $z_t$ given the observation history $o_{t}^H$, and $p(z_t)$ represents the prior distribution, modeled as a standard Gaussian distribution. All input observations are normalized to have a mean of 0 and variance of 1, aligning with the standard normal assumption for the prior.

%% file: Tables/01_Reward_table.tex
\begin{table}[t]
    \centering
    \caption{\textbf{Utilized reward terms}. 
    The value $g_{t}^{\text{PD}}$ is defined as $PD_{init} + \alpha_{\text{gain}} a_{t}^{\text{gain}}$, where $PD_{init}$ represents the initial $P_{init}$ and $D_{init}$ gains. In addition, $f^{\text{des}}_{\text{foot}, z}$ and $f_{\text{foot}, z}$ represent the desired and actual foot positions along the $z$-axis, respectively. $v_{xy}$ and $\omega_z$ denote the robot’s base horizontal velocity and angular velocity about the $z$-axis, respectively. Finally, $\tau$ and $\dot{q}_t$ represent the joint torque and joint angular velocity.}
    \vspace{-0.1in}
    \begin{center}
    \resizebox{0.99\linewidth}{!}
    {
        \def\arraystretch{1.5}
        \footnotesize
        \begin{tabular}{c|c|c} 
        \toprule
        \textbf{Reward}                                                         & \textbf{Equation}                   & \textbf{Weight}                       \\ \hline
        X, Y velocity tracking & $\exp\left(-4|v_{xy}^{cmd} - v_{xy}|^2\right)$ & 3.0 \\
        \hline
        Yaw velocity tracking & $\exp\left(-4|\omega_{z}^{cmd} - \omega_{z}|^2\right)$ & 1.5 \\
        \hline
        Z velocity  & $ v_{z}^2$ & -2.0 \\
        \hline
        Angular velocity & $ \omega_{xy}^2$ & -0.05 \\
        \hline
        Joint power & $  |\tau||\dot{q_t}|$ & -0.0001 \\
        \hline
        Action rate & $(a_{t-1} - a_t)^2$ & -0.01 \\ \hline
        Body height & $(h_\text{des} - h)^2$ & -10. \\ \hline
        Foot clearance & $(f^{des}_{\text{foot}, z} - f_{\text{foot}, z})^2$ & -0.4 \\ \hline  
        P gain limit           & $\exp\left(-|q_{\text{des}} - q_{t}|^2\right)$ & 0.25   \\ \hline
        Minimizing gain change    & $|g_{t-1}^{\text{PD}} - g_{t}^{\text{PD}}|^2$ & -0.01   \\ 
        
        \bottomrule
        \end{tabular} 

    }
    \end{center}
    \label{table:Reward_table}
    \vspace{-0.1in}
\end{table}

%% file: Sections/04_Experimental_Results.tex
\section{Experimental results}
\label{sec:experiments}
\input{Tables/02_Parameters}
Our experiments are organized into four main sections to validate the performance of the proposed GainAdaptor framework comprehensively.

First, simulation experiments (\secref{sec:Sim}) include an ablation study comparing the baseline algorithm~\cite{dreamwaq} with the final GainAdaptor model. This analysis highlights the performance improvements achieved by the proposed framework.

Second, indoor experiments (\secref{sec:In}) were conducted using the Unitree Go1 robot in a controlled test bed. These tests visualize the algorithm’s effectiveness and compare its performance against a widely used open-source algorithm, demonstrating its practical competitiveness.

Finally, outdoor experiments (\secref{sec:Out}) assessed the robustness of the framework in real-world scenarios. These included navigating a diverse long-range path and performing a zero-shot rock climbing task. The long-range navigation test demonstrated adaptability and stability across various terrains, while the zero-shot task evaluated the framework's flexibility in untrained and unstructured environments.

All training processes were conducted on a desktop PC with an Intel Core i5-12600KF CPU, 32GB RAM, and an NVIDIA RTX 3080 GPU. Details of our framework's hyperparameters are provided in~\Tabref{tab:ppo_hyperparameters}.

\input{Images/03_SIM_PD}

\subsection{Simulation Setting and Results}
\label{sec:Sim}
\subsubsection{PD Gain Adaptation Analysis}

The GainAdaptor framework optimizes PD gains dynamically to adapt to varying terrain conditions. ~\figref{fig:PD} illustrates the progression of P and D gains during the training process. The results show that the framework effectively adjusts the gains to maintain stability and improve terrain adaptability. Notably, the P gain is optimized to decrease to improve energy efficiency, while the D gain is strategically increased to enhance stability on challenging terrains, contributing to both stability and energy efficiency. This behavior aligns with the intended functionality of the GainAdaptor framework, ensuring robust locomotion across diverse environments.

\subsubsection{Terrain Classification Performance}
The terrain classification model plays a critical role in enabling the GainAdaptor framework to identify terrain types and adjust control parameters accordingly. The performance of the model is evaluated using multiclass classification metrics. ~\Tabref{tab:terrain_classification} summarizes the accuracy, precision, recall, and F1-score for each terrain type, including flat ground, slopes, rough terrain, and stairs. The results demonstrate high classification accuracy across all terrain types, confirming the model's reliability as a key component of the framework.

\input{Tables/03_2_Terrain}
\input{Tables/03_sim}

\subsubsection{Ablation Study}
To evaluate the contribution of individual components within the GainAdaptor framework, we conducted an ablation study in the Isaac Gym simulation environment~\cite{makoviychuk2021isaac}. This study compared the performance of the proposed method with its ablated variants by selectively removing key components, such as the dual-actor algorithm and the terrain classification module.

The baseline algorithm used for comparison is implemented based on the approach proposed in~\cite{dreamwaq}, closely following its methodology, and serves as the foundation for the GainAdaptor framework. The first variant, 'SA,' uses a single actor without the gain actor, while the second variant, 'NC,' includes the gain actor but removes the terrain classification module.

The performance of each method was evaluated based on the walking speed tracking accuracy and energy efficiency. Tracking accuracy was assessed using the RMSE, and energy efficiency was analyzed through the calculation of average torque and power. The average power $P$ was computed as follows:

\begin{equation}
P = \frac{1}{12} \sum_{i=1}^{12} | \tau_i \cdot \dot{q}_i |
\label{Equ:Average}
\end{equation}

where $\tau_i$ represents the torque of the $i$-th motor, and $\dot{q}_i$ denotes its angular velocity.

The results, presented in~\Tabref{tab:sim_table}, highlight the significant impact of the key components on terrain adaptability and energy efficiency. The inclusion of the gain actor, as demonstrated in the NC variant, effectively reduces torque and power consumption, showcasing its contribution to energy efficiency. Furthermore, adding the terrain classification module to NC leads to a notable decrease in RMSE for walking speed tracking, emphasizing its critical role in enhancing tracking accuracy. These findings underline the complementary nature of the gain actor and terrain classification module, validating the GainAdaptor framework’s holistic design in achieving robust and efficient quadrupedal locomotion.

\input{Tables/04_Real}
\input{Images/04_INDOOR}
\subsection{Test Bed Results}

The test bed experiments were conducted in a controlled indoor environment to validate the real-world performance of the GainAdaptor framework. Using the Go1 robot, the framework's performance was compared against two benchmarks: the baseline algorithm~\cite{dreamwaq}, and the WTW algorithm~\cite{margolis2023walk}, a widely used open-source approach for quadrupedal locomotion. The comparison utilized evaluation metrics such as P gain, D gain, torque, and power consumption to comprehensively assess the performance.

The GainAdaptor framework consistently demonstrated superior performance across all experimental scenarios. On average, power consumption was reduced by 56.71\% compared to the baseline~\cite{dreamwaq} and 33.07\% compared to WTW~\cite{margolis2023walk}, effectively improving energy efficiency through optimized control gains. However, in terms of torque, the reduction compared to WTW~\cite{margolis2023walk} was limited or showed no significant difference. This outcome is attributed to the initial P gain settings: WTW~\cite{margolis2023walk} started with a relatively lower P gain of 20, while the baseline~\cite{dreamwaq} and GainAdaptor framework started with a higher initial P gain of 28. A higher initial P gain leads to relatively higher initial torque requirements, as reflected in the experimental results. Adjusting the initial P gain to a lower value and allowing the framework to further reduce P gain during the learning process is expected to achieve additional torque reduction, highlighting the potential for further improvement in the GainAdaptor framework and contributing to greater energy efficiency.

Additionally, the GainAdaptor framework recorded lower power consumption compared to the baseline~\cite{dreamwaq} and WTW algorithms~\cite{margolis2023walk} on irregular and dynamic terrains, such as stairs, obstacles, and seesaws. These results indicate that the GainAdaptor framework enables the robot to operate reliably and stably in unpredictable and diverse environments.

In particular, as shown in~\figref{fig:torquevar}, the torque distribution of the GainAdaptor framework is significantly more uniform compared to the baseline~\cite{dreamwaq} and WTW algorithms~\cite{margolis2023walk}. To quantify this, we calculated the variance of the torque values across all joints: the baseline~\cite{dreamwaq} and WTW algorithms~\cite{margolis2023walk} exhibit variances of 2.8180 and 4.9126, respectively, while the GainAdaptor framework achieves a much lower variance of 0.7868. This represents a 72.08\% reduction in variance compared to the baseline~\cite{dreamwaq} and an 83.98\% reduction compared to WTW~\cite{margolis2023walk}.

The significant decrease in torque variance demonstrates the GainAdaptor framework's ability to distribute torque evenly across joints, reducing mechanical stress and mitigating motor overheating issues caused by prolonged high torque. This enhancement improves locomotion stability, long-range reliability, and adaptability to diverse environments, while lowering the risk of motor failure. Overall, the test bed results confirm that the GainAdaptor framework offers an efficient and stable control strategy across various terrains. Refining initial settings, such as the P gain, could further enhance its contributions to energy efficiency and robust control.

\label{sec:In}
\subsection{Outdoor Results}
\input{Images/04_Outdoor}

\input{Images/05_zeroshot}
\label{sec:Out}
\subsubsection{Long-range Navigation}
The long-range navigation experiment involved the robot traversing a predefined outdoor course as shown in~\figref{fig:longterm}. The course consisted of multiple segments featuring varying terrains, including grass (A), stairs with leaf debris (B-1, B-2), forest paths covered with dry leaves (C), a flat playground area with small obstacles (D), and sloped grassy surfaces (E). These terrains were chosen to reflect real-world challenges such as uneven surfaces, dynamic transitions, and debris. The GainAdaptor framework enabled the robot to maintain stable and energy-efficient locomotion throughout the entire course. The framework successfully adapted to diverse terrains by dynamically adjusting the PD gains based on terrain conditions. Notably, the robot completed the course without significant failures, showcasing its ability to navigate unpredictable and complex environments.

\subsubsection{Zero-Shot Rock Climbing}
In addition to the predefined course, a zero-shot task was performed to assess the framework’s capability to handle scenarios outside the scope of its training data. The robot was tasked with climbing a rock higher than its body height as shown in~\figref{fig:zeroshot}. Despite no prior training for such a task, the GainAdaptor framework enabled the robot to achieve successful rock climbing by dynamically modifying control parameters in real-time. This experiment demonstrates the inherent adaptability of the GainAdaptor framework. By optimizing both P and D gains, the robot could exert sufficient force to climb the obstacle while maintaining stability. The success of this task underscores the framework’s potential for applications in challenging and unstructured environments where zero-shot adaptability is essential.

\label{sec:zeroshot}

%% file: Tables/02_Parameters.tex
\begin{table}[t]
    \centering
    \caption{\textbf{Hyperparameters for training}}
    \label{tab:ppo_hyperparameters}
    \begin{tabular}{l|c}
        \hline
        \textbf{Hyperparameter}                  & \textbf{Value}        \\ 
        \hline
        Learning Rate                            & $10^{-3}$             \\ 
        PPO Clip Range                           & 0.2                   \\ 
        Discount Factor (Gamma)                  & 0.99                  \\ 
        GAE Lambda                               & 0.95                  \\ 
        Dual Actors Network Hidden Layers              & [512, 256, 128]       \\ 
        Value Network Hidden Layers              & [512, 256, 128]       \\ 
        Estimator Network Hidden Layers          & [256, 128]            \\ 
        Classifier Network Hidden Layers          & [256, 128]            \\ 
        Entropy Coefficient                      & 0.01                  \\ 
        Value Loss Coefficient                   & 1.0                   \\ 
        Optimizer                                & Adam                  \\ 
        \hline
    \end{tabular}

\vspace{-0.1in}
    
\end{table}

%% file: Images/03_SIM_PD.tex
\begin{figure*}[t]
\centering 
\begin{tabular}{c@{\hskip 0.001\linewidth}c@{\hskip 0.001\linewidth}c}
\includegraphics[width=0.9\linewidth]{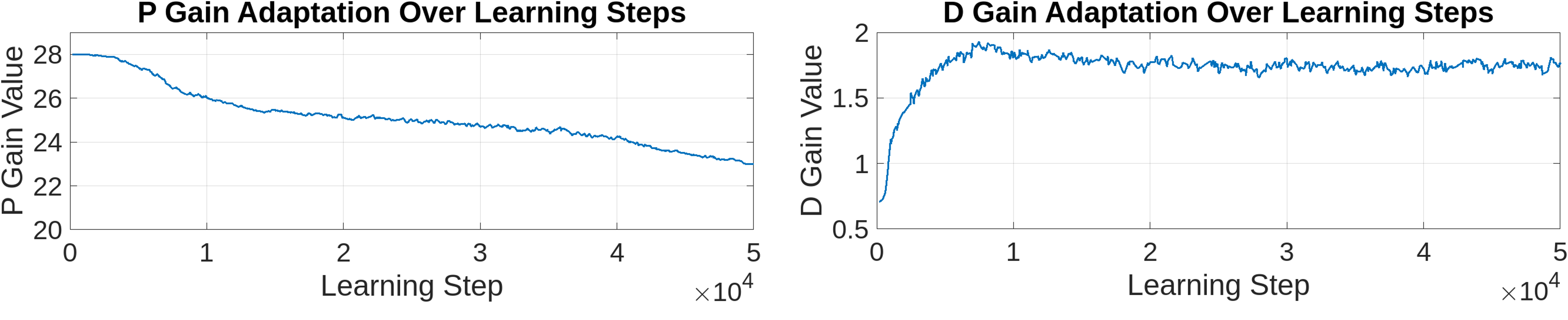} & 

\end{tabular}
\caption{{\bf Progression of P and D gains during the training process.} The GainAdaptor framework dynamically adjusts the P and D gains based on the terrain conditions.}

\label{fig:PD} 
\vspace{-0.1in}

\end{figure*}

%% file: Tables/03_2_Terrain.tex
\begin{table}[t]
\centering
\caption{\textbf{Performance metrics of terrain classification model}}
\vspace{-0.1in}
\begin{center}
\resizebox{1.0\linewidth}{!}{%
    \renewcommand{\arraystretch}{1.2} 
    \begin{tabular}{c|c|c|c|c}
    \hline
    \textbf{Terrain Type} & \textbf{Accuracy (\%)} & \textbf{Precision (\%)} & \textbf{Recall (\%)} & \textbf{F1-Score (\%)} \\ \hline
    Level Ground           & 95.2                  & 94.8                   & 95.0                & 94.9                   \\ 
    Slopes                & 92.7                  & 91.5                   & 92.3                & 91.9                   \\ 
    Rough Terrain         & 90.8                  & 89.6                   & 90.2                & 89.9                   \\ 
    Stairs                & 88.5                  & 87.4                   & 88.1                & 87.7                   \\ \hline
    \cellcolor{Gray1}\textbf{Overall Avg} & \cellcolor{Gray1}\textbf{91.8}         & \cellcolor{Gray1}\textbf{90.8}          & \cellcolor{Gray1}\textbf{91.4}       & \cellcolor{Gray1}\textbf{91.1}          \\ \hline
    \end{tabular}%
}
\end{center}
\vspace{-0.1in}
\label{tab:terrain_classification}
\end{table}

%% file: Tables/03_sim.tex
\begin{table}[t]
\centering
\caption{\textbf{Performance comparison in Isaac Gym simulation across different speeds and metrics.}
The table presents a comparative analysis using metrics such as Root-Mean-Square Error(RMSE) between command velocity and base velocity, Mean Torque, and Mean Power at base speeds of 1.0 m/s and 2.0 m/s. Best performances are highlighted in \textbf{bold}.
}
\vspace{-0.1in}
\begin{center}
\resizebox{1.0\linewidth}{!}
{
    \def\arraystretch{1.00}
    \footnotesize
    \begin{tabular}{c|c|cccc} 
    \toprule
    \multirow{2}{*}{\textbf{Speed (m/s)}} & {\multirow{2}{*}{\shortstack[alignment]{\textbf{Metric}}}} & \multicolumn{4}{c}{\textbf{Methods}} \\ \cline{3-6}
    &  & Baseline~\cite{dreamwaq} & SA & NC & Ours  \\ \hline

\multirow{3}{*}{\shortstack[alignment]{1.0}}  & RMSE   & 0.1068 & 0.2223 & 0.1677 & \textbf{0.0782}\\ 
            & Torque(Nm)  & 3.6856 & 3.0554 & 2.8549 & \textbf{2.6876}    \\
            & Power(W) & 7.9383  & 8.1064 & 4.9225 & \textbf{3.2813}      \\ \hline
\multirow{3}{*}{\shortstack[alignment]{2.0}}  & RMSE & 0.2818 & 0.4347 & 0.2223 & \textbf{0.1975}\\ 
            & Torque(Nm)  & 4.2298
 & 4.0997 & 3.8447 & \textbf{3.6754}    \\
            & Power(W) & 17.9411  & 13.5053 & 11.7928 &  \textbf{10.3694}      \\ \hline

    \bottomrule
    \end{tabular}

}

\end{center}
\vspace{-0.1in}
\label{tab:sim_table}
\end{table}

%% file: Tables/04_Real.tex
\begin{table*}[t]
\centering
\caption{\textbf{Performance comparison across various terrains.} This table presents a detailed comparison of performance metrics for the proposed GainAdaptor framework against the baseline\cite{dreamwaq} and WTW algorithms\cite{margolis2023walk} across six different terrains: Stairs (A), Obstacles (B), Seesaw (C), Rough Terrain (D), Slopes (E), and Plain Ground (F). Metrics include P gain, D gain, torque, and power consumption, where lower torque and power indicate better performance.}
\vspace{-0.1in}
\begin{center}
\resizebox{0.98\linewidth}{!}{%
    \def\arraystretch{0.95}
    \footnotesize
    \begin{tabular}{c|c|cccc|cccc|cccc}
        \toprule
        \multirow{2}{*}{\textbf{Method}} & \multirow{2}{*}{\textbf{Leg}} & \multicolumn{4}{c|}{\textbf{Stairs (A)}} & \multicolumn{4}{c|}{\textbf{Obstacle (B)}}
        & \multicolumn{4}{c}{\textbf{Seesaw (C)}}
        \\ \cline{3-14}
        &  & P Gain & D Gain & Torque $\downarrow$ & Power $\downarrow$ & P Gain & D Gain & Torque $\downarrow$ & Power $\downarrow$ & P Gain & D Gain & Torque $\downarrow$ & Power $\downarrow$  \\
        \hline
        \multirow{5}{*}{Baseline\cite{dreamwaq}} & Front Left & 28 & 0.7 & 4.6467 & 7.0407 & 28 & 0.7 & 4.5663 & 4.4817 & 28 & 0.7 & 4.8272 & 5.3321 \\
        & Front Right & 28 & 0.7 & 5.0839 & 10.8278 & 28 & 0.7 & 5.1205 & 7.4182 & 28 & 0.7 & 4.7958 & 8.5984 \\
        & Rear Left & 28 & 0.7 & 3.0879 & 6.2626 & 28 & 0.7 & 2.9987 & 4.6431 & 28 & 0.7 & 3.0334 & 5.1551 \\
        & Rear Right & 28 & 0.7 & 3.1038 & 6.3028 & 28 & 0.7 & 2.3853 & 3.8913 & 28 & 0.7 & 3.1942 & 4.6195 \\
        
        & \multicolumn{1}{c|}{\cellcolor{Gray1}Avg} & \cellcolor{Gray1}28 & \cellcolor{Gray1}0.7 & \cellcolor{Gray1}3.9806 & \cellcolor{Gray1}7.6085 & \cellcolor{Gray1}28 & \cellcolor{Gray1}0.7 & \cellcolor{Gray1}3.7677 & \cellcolor{Gray1}5.1086 &
        \cellcolor{Gray1}28 & \cellcolor{Gray1}0.7 & \cellcolor{Gray1}3.9627 & \cellcolor{Gray1}5.9263 \\
        \hline
        \multirow{5}{*}{WTW\cite{margolis2023walk}} & Front Left & 20 & 0.5 & 3.4987 & 4.9081 & 20 & 0.5 & 3.1991 & 3.6887 & 20 & 0.5 & 4.1552 & 5.2671 \\
        & Front Right & 20 & 0.5 & 3.0022 & 4.2648 & 20 & 0.5 & 3.0925 & 4.1903 & 20 & 0.5 & 4.4741 & 5.9888 \\
        & Rear Left & 20 & 0.5 & 3.7076 & 6.1636 & 20 & 0.5 & 3.4470 & 5.0872 & 20 & 0.5 & 3.7584 & 3.3701 \\
        & Rear Right & 20 & 0.5 & 2.9866 & 4.3501 & 20 & 0.5 & 3.1422 & 3.8433 & 20 & 0.5 & 3.4764 & 2.8129 \\
        
        & \multicolumn{1}{c|}{\cellcolor{Gray1}Avg} & \cellcolor{Gray1}20 & \cellcolor{Gray1}0.5 & \cellcolor{Gray1}3.2988 & \cellcolor{Gray1}4.9217 & \cellcolor{Gray1}20 & \cellcolor{Gray1}0.5 & \cellcolor{Gray1}3.2202 & \cellcolor{Gray1}4.2024 &
        \cellcolor{Gray1}20 & \cellcolor{Gray1}0.5 & \cellcolor{Gray1}3.9660 & \cellcolor{Gray1}4.3597 \\
        \hline
        
        \multirow{5}{*}{GainAdaptor} & Front Left & 23.028 & 1.6646 & 4.0113 & 3.4858 & 23.0217 & 1.8272 & 4.1167 & 4.0970 & 23.0377 & 1.6398 & 3.7819 & 3.9166 \\
        & Front Right & 23.028 & 1.8071 & 3.8616 & 4.26 & 23.0193 & 1.7824 & 3.6547 & 3.7998 & 23.0387 & 1.7843 & 4.4471 & 4.4779 \\
        & Rear Left & 23.029 & 1.6346 & 3.4866 & 2.1815 & 23.0227 & 1.5978 & 2.6674 & 1.7231 & 23.04 & 1.6164 & 3.4743 & 2.4124 \\
        & Rear Right & 23.0267 & 1.8447 & 3.8604 & 2.3676 & 23.0177 & 1.6957 & 3.9767 & 3.2012 & 23.0347 & 1.8141 & 3.3718 & 2.2710 \\
        & \multicolumn{1}{c|}{\cellcolor{Gray1}Avg} & \cellcolor{Gray1}23.0279 & \cellcolor{Gray1}1.7378 & \cellcolor{Gray1}3.8050 & \cellcolor{Gray1}3.0737 & \cellcolor{Gray1}23.0204 & \cellcolor{Gray1}1.7258 & \cellcolor{Gray1}3.6039 & \cellcolor{Gray1}3.2053 &
        \cellcolor{Gray1}23.0377 & \cellcolor{Gray1} 1.7137 & \cellcolor{Gray1}3.7688 & \cellcolor{Gray1}3.2695 \\
        \bottomrule

        \toprule
        \multirow{2}{*}{\textbf{Method}} & \multirow{2}{*}{\textbf{Leg}} & \multicolumn{4}{c|}{\textbf{Rough (D)}} & \multicolumn{4}{c|}{\textbf{Slope (E)}}
        & \multicolumn{4}{c}{\textbf{Plain (F)}}
        \\ \cline{3-14}
        &  & P Gain & D Gain & Torque $\downarrow$ & Power $\downarrow$ & P Gain & D Gain & Torque $\downarrow$ & Power $\downarrow$ & P Gain & D Gain & Torque $\downarrow$ & Power $\downarrow$  \\
        \hline
        \multirow{5}{*}{Baseline\cite{dreamwaq}} & Front Left & 28 & 0.7 & 5.1880 & 5.6696 & 28 & 0.7 & 4.4944 & 5.6963 & 28 & 0.7 & 4.1533 & 3.9502 \\
        & Front Right & 28 & 0.7 & 4.7530 & 7.0304 & 28 & 0.7 & 5.0957 & 9.5448 & 28 & 0.7 & 4.5134 & 6.2647 \\
        & Rear Left & 28 & 0.7 & 3.2619 & 5.0877 & 28 & 0.7 & 3.1806 & 5.2475 & 28 & 0.7 & 2.9605 & 4.0861 \\
        & Rear Right & 28 & 0.7 & 3.0995 & 3.7266 & 28 & 0.7 & 3.0714 & 5.0740 & 28 & 0.7 & 2.7494 & 3.7271 \\
        & \multicolumn{1}{c|}{\cellcolor{Gray1}Avg} & \cellcolor{Gray1}28 & \cellcolor{Gray1}0.7 & \cellcolor{Gray1}4.0756 & \cellcolor{Gray1}5.3786 & \cellcolor{Gray1}28 & \cellcolor{Gray1}0.7 & \cellcolor{Gray1}3.9605 & \cellcolor{Gray1}6.3907 &
        \cellcolor{Gray1}28 & \cellcolor{Gray1}0.7 & \cellcolor{Gray1}3.5941 & \cellcolor{Gray1}4.5070 \\
        \hline
        \multirow{5}{*}{WTW\cite{margolis2023walk}} & Front Left & 20 & 0.5 & 3.3446 & 3.4060 & 20 & 0.5 & 3.6058 & 4.7971 & 20 & 0.5 & 3.1103 & 3.2329 \\
        & Front Right & 20 & 0.5 & 3.2779 & 4.4367 & 20 & 0.5 & 3.7384 & 4.6571 & 20 & 0.5 & 3.0428 & 3.5574 \\
        & Rear Left & 20 & 0.5 & 3.3096 & 5.0580 & 20 & 0.5 & 3.8769 & 5.5211 & 20 & 0.5 & 3.1951 & 4.7292 \\
        & Rear Right & 20 & 0.5 & 3.3260 & 4.4090 & 20 & 0.5 & 3.7618 & 5.8094 & 20 & 0.5 & 2.9310 & 3.4794 \\
        
        & \multicolumn{1}{c|}{\cellcolor{Gray1}Avg} & \cellcolor{Gray1}20 & \cellcolor{Gray1}0.5 & \cellcolor{Gray1}3.3145 & \cellcolor{Gray1}4.3274 & \cellcolor{Gray1}20 & \cellcolor{Gray1}0.5 & \cellcolor{Gray1}3.7457 & \cellcolor{Gray1}5.1962 &
        \cellcolor{Gray1}20 & \cellcolor{Gray1}0.5 & \cellcolor{Gray1}3.0698 & \cellcolor{Gray1}3.7497 \\
        \hline
        
        \multirow{5}{*}{GainAdaptor} & Front Left & 23.0673 & 1.6208 & 4.0457 & 3.8392 & 23.0027 & 1.6531 & 4.2025 & 3.2957 & 23.0703 & 1.6441 & 3.8667 & 3.6469 \\
        & Front Right & 23.0683 & 1.7741 & 3.8148 & 4.3448 & 23.003 & 1.7969 & 3.5925 & 3.9436 & 23.071 & 1.7931 & 3.7628 & 4.0667 \\
        & Rear Left & 23.068 & 1.6111 & 3.3765 & 2.7159 & 23.0033 & 1.6091 & 3.2045 & 2.4952 & 23.0723 & 1.6332 & 3.2294 & 1.9181 \\
        & Rear Right & 23.067 & 1.8323 & 3.6786 & 2.2757 & 23.0017 & 1.8362 & 3.9315 & 2.3138 & 23.069 & 1.8305 & 3.3621 & 1.7728 \\
        & \multicolumn{1}{c|}{\cellcolor{Gray1}Avg} & \cellcolor{Gray1}23.0677 & \cellcolor{Gray1}1.7096 & \cellcolor{Gray1}3.7289 & \cellcolor{Gray1}3.2939 & \cellcolor{Gray1}23.0027 & \cellcolor{Gray1}1.7238 & \cellcolor{Gray1}3.7328 & \cellcolor{Gray1}3.0121 &
        \cellcolor{Gray1}23.0707 & \cellcolor{Gray1}1.7252 & \cellcolor{Gray1}3.5552 & \cellcolor{Gray1}2.8511 \\
        \bottomrule

    \end{tabular}%
}
\end{center}
\label{table:indoor}
\vspace{-0.2in}
\end{table*}

%% file: Images/04_INDOOR.tex
\begin{figure}[t!]
\centering 
\begin{tabular}{c}
\includegraphics[width=0.98\linewidth]{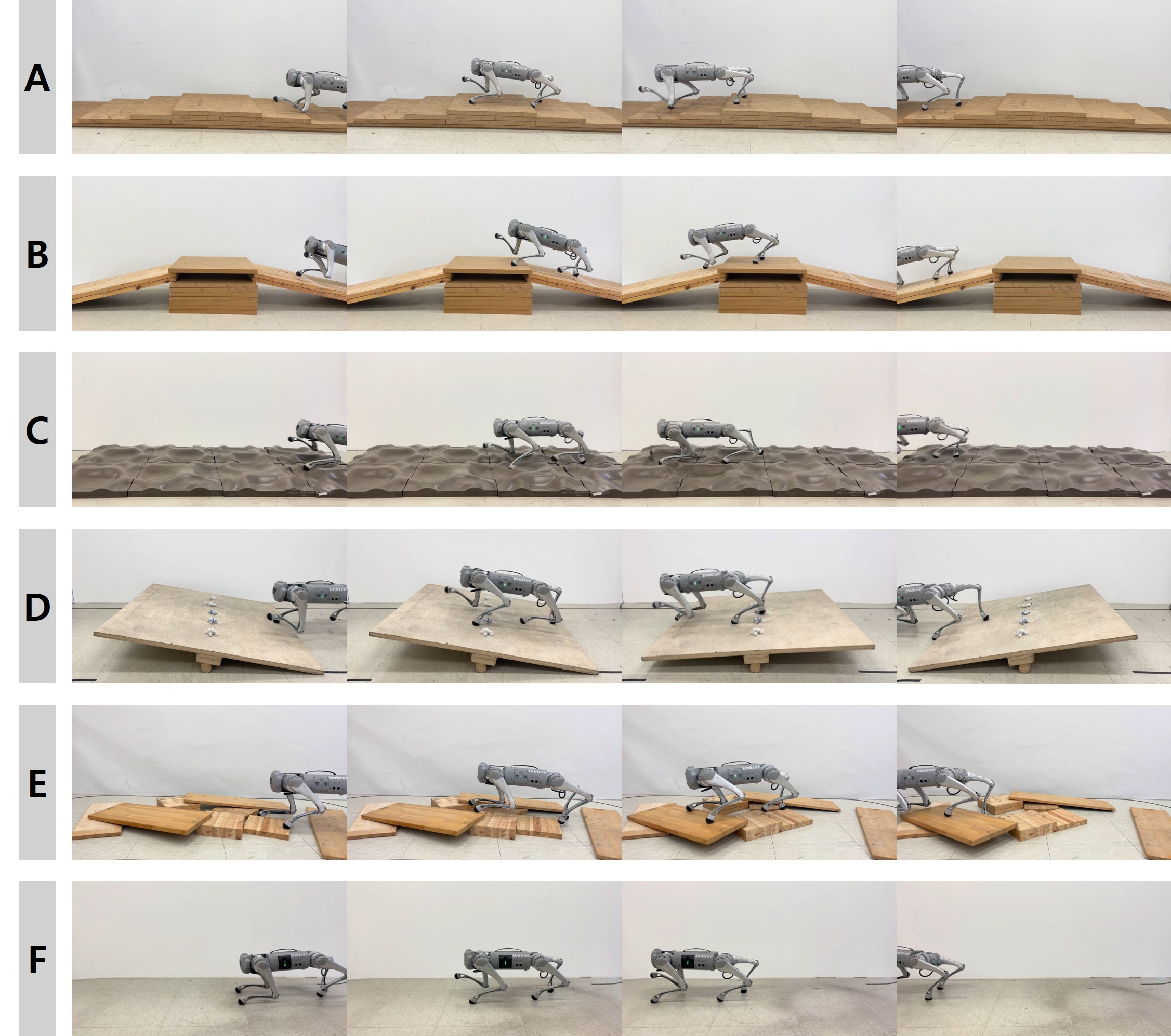} \vspace{-0.05in} \\ 
\end{tabular}

\caption{{\bf Indoor experiments on various terrains.}
}
\vspace{-0.1in}
\label{fig:indoor} 
\end{figure}

\begin{figure}[t!]
\centering 
\begin{tabular}{c}
\includegraphics[width=0.98\linewidth]{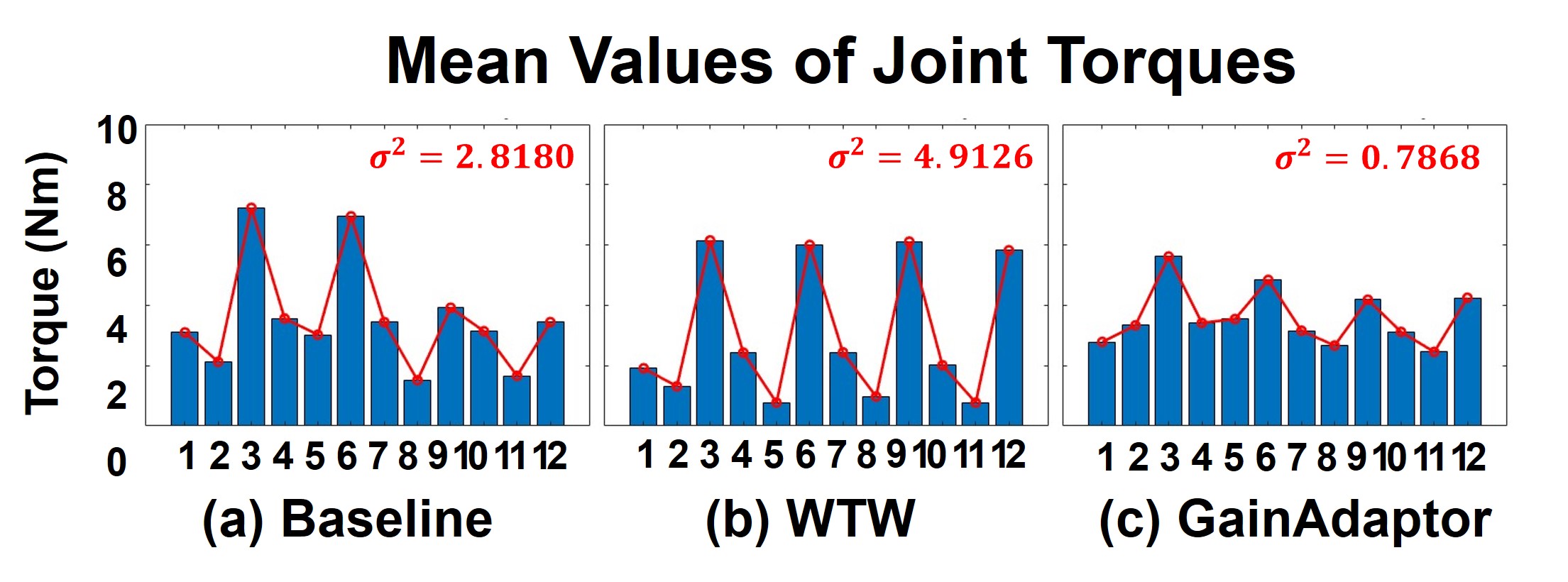} \vspace{-0.05in} \\ 
\end{tabular}

\caption{{\bf Comparison of mean joint torque on level ground.}}
\vspace{-0.1in}
\label{fig:torquevar} 
\end{figure}

%% file: Images/04_Outdoor.tex
\begin{figure*}[t]
\begin{center}
{
\begin{tabular}{c}
\includegraphics[width=0.99\linewidth]{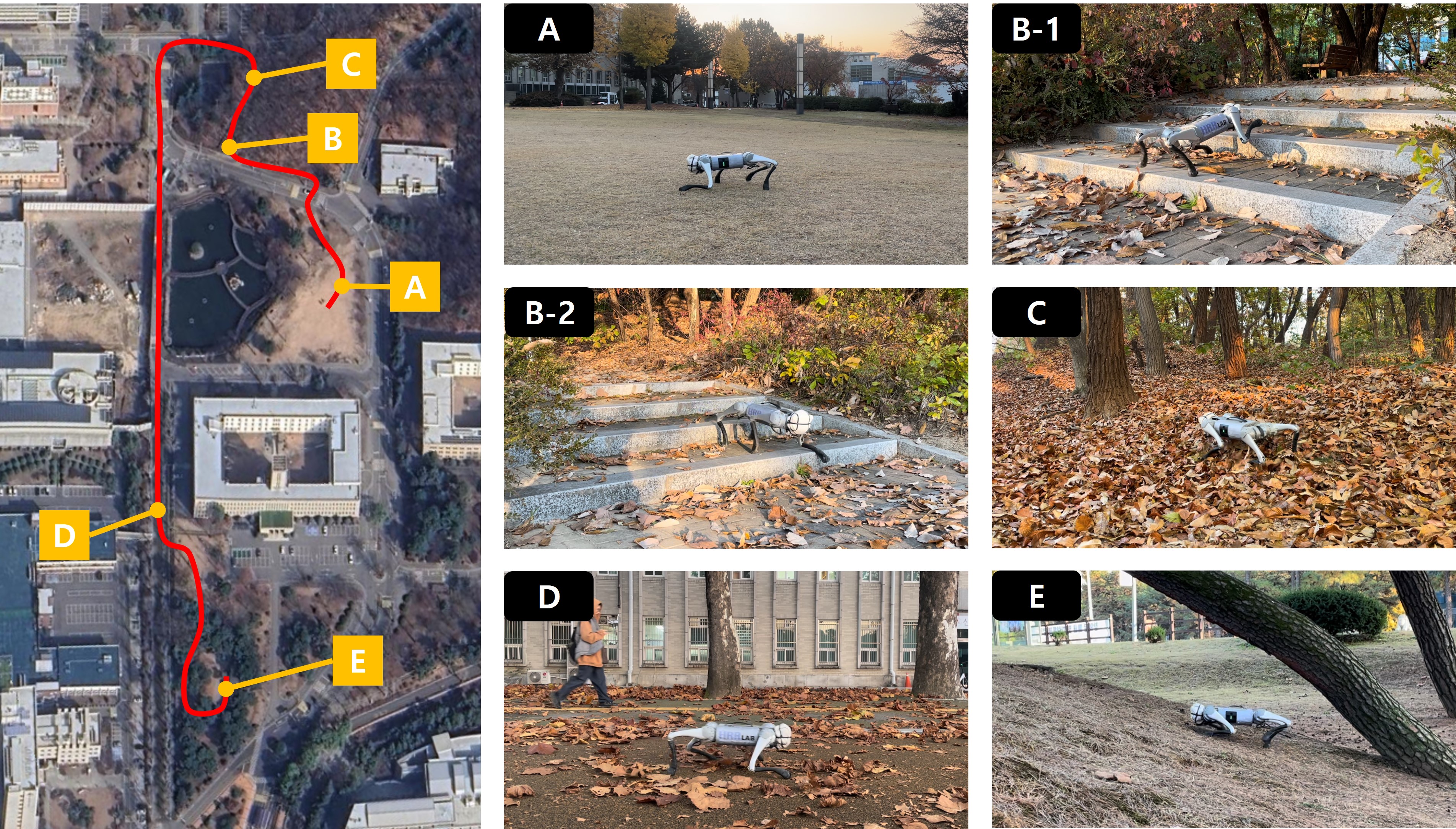} \\

\end{tabular}
}
\end{center}
\caption{{\bf Long-range navigation course overview and representative terrains.} The left image shows the aerial view of the long-range navigation course marked in red, with segments labeled A through E. The right images show representative examples of the robot traversing different terrains.}

\label{fig:longterm} 
\vspace{-0.15in}

\end{figure*}

%% file: Images/05_zeroshot.tex
\begin{figure}[t]
\centering 
\begin{tabular}{c}
\includegraphics[width=0.98\linewidth]{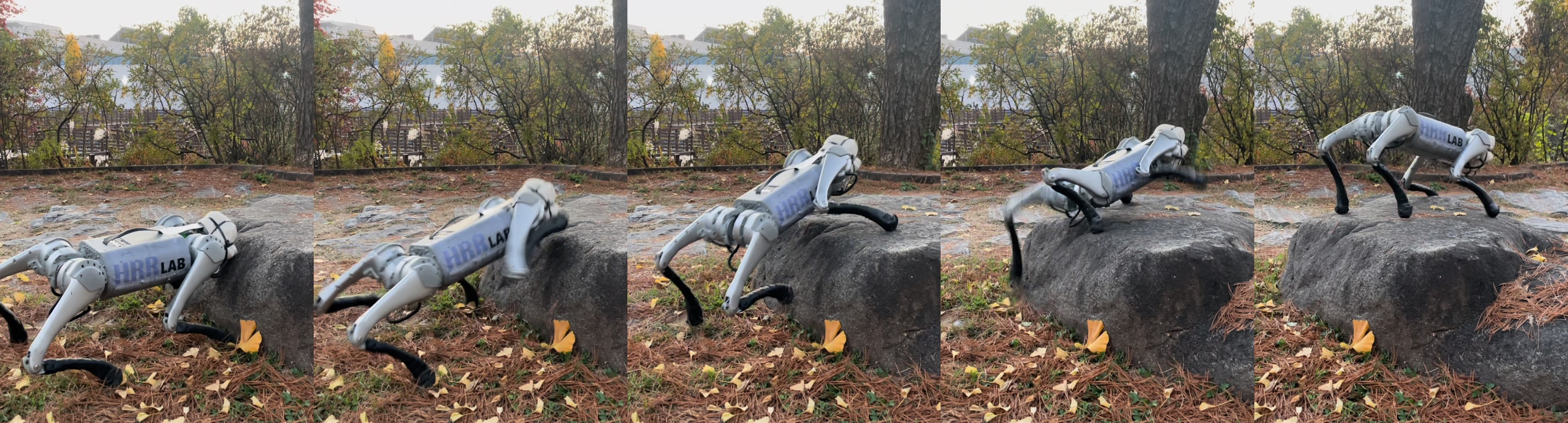}

\end{tabular}
\caption{{\bf Zero-shot rock climbing sequence.} The sequence of images shows the robot attempting to climb a rock higher than its body height, despite not being explicitly trained for such a task. The progression demonstrates the GainAdaptor framework’s adaptability.}

\label{fig:zeroshot} 
\vspace{-0.1in}

\end{figure}

%% file: Sections/05_Conclusion.tex
\section{Conclusion}

In this paper, we proposed the GainAdaptor framework to implement energy-efficient quadrupedal locomotion for robots. We utilized a dual-actor approach to maintain performance consistency even in environments with different labels, such as PD control gains and target joint angles, which significantly improved performance while increasing model efficiency and simplifying training in complex multi-label environments.

As a result of experiments, the GainAdaptor framework achieved significantly higher energy efficiency, reducing power consumption by up to \textbf{33.07\%} and torque variance by \textbf{83.98\%} compared to WTW~\cite{margolis2023walk}. Additionally, it exhibited exceptional adaptability in complex terrains and zero-shot scenarios. These findings highlight the GainAdaptor framework's potential to enhance energy efficiency, reduce mechanical stress, and ensure robust performance in real-world applications.

In future research, we aim to expand optimization to visual locomotion for dynamic environments. 